\documentclass[final]{l4dc2026}
\usepackage{zshi}
\usepackage{enumitem}
\usepackage{subcaption}
\usepackage{makecell}
\usepackage{hyperref}
\usepackage{zref-clever}
\newcommand{\cref}[1]{\zcref{#1}}
\newcommand{\Cref}[1]{\zcref[S]{#1}}

\title{Certified Training with Branch-and-Bound\\ \ for Lyapunov-stable Neural Control}
\usepackage{times}

\author{%
 \Name{Zhouxing Shi} \Email{zhouxing.shi@ucr.edu}\\
 \addr University of California, Riverside, CA 92521
 \AND
 \Name{Haoyu Li} \Email{haoyuli5@illinois.edu}\\
 \addr University of Illinois Urbana-Champaign, IL 61801
 \AND
 \Name{Cho-Jui Hsieh} \Email{chohsieh@cs.ucla.edu}\\
 \addr University of California, Los Angeles, CA 90095
 \AND
 \Name{Huan Zhang} \Email{huan@huan-zhang.com}\\
 \addr University of Illinois Urbana-Champaign, IL 61801
}

\newif\ifcomment
\commenttrue

\ifcomment
    \newcommand{\huan}[1]{\textcolor{blue}{Huan: #1}}
\else
    \newcommand{\huan}[1]{}
\fi

\begin{document}

\maketitle

\begin{abstract}
We study the problem of learning verifiably Lyapunov-stable neural controllers that provably satisfy the Lyapunov asymptotic stability condition within a region-of-attraction (ROA). Unlike previous works that adopted counterexample-guided training without considering the computation of verification in training, we introduce Certified Training with Branch-and-Bound (CT-BaB), a new certified training framework that optimizes certified bounds, thereby reducing the discrepancy between training and test-time verification that also computes certified bounds. To achieve a relatively global guarantee on an entire input region-of-interest, we propose a training-time BaB technique that maintains a dynamic training dataset and adaptively splits hard input subregions into smaller ones, to tighten certified bounds and ease the training. Meanwhile, subregions created by the training-time BaB also inform test-time verification, for a more efficient training-aware verification. We demonstrate that CT-BaB yields verification-friendly models that can be more efficiently verified at test time while achieving stronger verifiable guarantees with larger ROA. On the largest output-feedback 2D Quadrotor system experimented, CT-BaB reduces verification time by over 11$\times$ relative to the previous state-of-the-art baseline using Counterexample Guided Inductive Synthesis (CEGIS),  while achieving 164$\times$ larger ROA. Code is available at \url{https://github.com/shizhouxing/CT-BaB}.

\end{abstract}

\begin{keywords}
Lyapunov stability, certified training, neural network verification, branch-and-bound
\end{keywords}

\section{Introduction}

Deep learning with neural networks (NNs) has significantly advanced many domains, including control and robotic systems, where NN-based controllers are increasingly adopted. Despite their impressive capability, it remains challenging to obtain certified guarantees on the behaviors of NNs. However, these guarantees are critical for the trustworthy deployment of NNs in mission-critical domains~\citep{ames2016control,bansal2017hamilton,chang2019neural}. In this work, we focus on learning NN-based controllers that are \emph{formally verifiable} to be Lyapunov asymptotically stable in discrete-time nonlinear dynamical systems, which is usually approached by training the controller jointly with a Lyapunov function~\citep{wu2023neural,yang2024lyapunov}. Intuitively, a Lyapunov function provides an energy-like measure on the system state: it is positive definite and attains its global minimum at the equilibrium. The Lyapunov condition~\citep{lyapunov1992general} then requires that, for every state within the region-of-attraction (ROA), the closed-loop update yields a strictly smaller Lyapunov function value, i.e., the system evolves toward lower ``energy.'' When these properties are verified, any trajectory initialized in the ROA is guaranteed to converge to the equilibrium, thereby establishing asymptotic stability.

Previous works~\citep{wu2023neural,yang2024lyapunov} on this task were commonly based on the Counterexample Guided Inductive Synthesis (CEGIS) framework by iteratively searching for counterexamples and training models (both controllers and Lyapunov functions) on the counterexamples. The Lyapunov condition is subsequently verified by a formal verifier~\citep{zhang2018efficient,xu2020automatic} for the trained models.
However, training on individual counterexamples does not consider the computation of verification required at test time (i.e., the offline formal verification after training is complete, which we call \emph{test-time verification}), and thus the models are often not ``verification-friendly'', leading to time-consuming verification. For instance, \citet{yang2024lyapunov} spent 8.9 hours to verify a controller for the 2D quadrotor system with output feedback. This difficulty also restricts the size of ROA that can be achieved.

In this paper, we propose a novel \textbf{C}ertified \textbf{T}raining framework enhanced with training-time \textbf{B}ranch-\textbf{a}nd-\textbf{B}ound, namely \textbf{CT-BaB}.
By ``\emph{certified training}''~\citep{gowal2018effectiveness,mirman2019provable,zhang2019towards,shi2021fast}, we optimize differentiable \emph{certified bounds} on violations given \emph{subregions} of inputs, rather than violations on \emph{individual counterexamples} in CEGIS. Since test-time verification is also achieved by computing certified bounds, the training is now \emph{verification-aware} to produce verification-friendly models.
Unlike existing certified training works that primarily focused on the \emph{local} robustness of NNs, we need {relatively global} guarantees on an entire input region-of-interest, which is too large for standard certified training to handle directly. To address this, we propose a \emph{training-time branch-and-bound} (BaB) attached to certified training. We maintain a  training dataset that consists of subregions within the region-of-interest, and we dynamically split hard subregions into smaller ones, which eases the training with tighter certified bounds.
Although verification with more extensive BaB is still conducted at test time to fully verify models, subregions created during our training can be exported to bootstrap and significantly speed up test-time verification, making the verification \emph{training-aware}. 

We demonstrate CT-BaB on learning asymptotically Lyapunov-stable NN controllers, and to the best of our knowledge, this is the first certified training approach for this task.
We show that CT-BaB produces models with much larger ROA, while enabling much faster test-time verification, thus producing verification-friendly models with stronger guarantees.
Notably, compared to the previous state-of-the-art~\citep{yang2024lyapunov}, on the largest 2D quadrotor system: for the state feedback setting, CT-BaB reduces the verification time from 1.1hrs to 49s (\textbf{80$\times$ faster}) while yielding a \textbf{16$\times$ larger ROA}; for the output feedback setting, CT-BaB reduces the verification time from 8.9hrs to 0.8hrs (\textbf{11$\times$ faster}) with a \textbf{164$\times$ larger ROA}.

\vspace{-10pt}
\section{Methodology}

\subsection{Background and Problem Settings}
\label{sec:problem}

\paragraph{Lyapunov-stable neural control.}
We focus on learning verifiably stable neural controllers with Lyapunov asymptotic stability guarantees~\citep{lyapunov1992general}, for a nonlinear discrete-time dynamical system with continuous control actions. We adopt the problem formulation from \citet{yang2024lyapunov}. Suppose the input state has $d$ dimensions, a nonlinear dynamical system is defined as
\begin{equation}
\rvx_{t+1}=f(\rvx_t,u(\rvx_t)),
\label{eq:system}
\end{equation}
where $\rvx_t$ is the state at the current time step $t$, a continuous control input $u(\rvx_t)\in\sR^{n_u}$ is generated by a NN-based controller, and the system dynamics determines the next state $\rvx_{t+1}$.
It is assumed that the system dynamics are known and there exists an equilibrium $\rvx^*$ such that $\rvx^*=f(\rvx^*,u(\rvx^*))$. The analysis focuses on a region-of-interest (ROI) $\gB\in\sR^d$ containing the equilibrium, i.e., $\rvx^*\in\gB$.
We assume $\gB$ is an axis-aligned bounding box $ \gB=\{ \rvx \mid \ul{\rvb}\!\leq\! \rvx \!\leq\! \ol{\rvb},\,\rvx\in\sR^d\}$ with boundary $\ul{\rvb}, \ol{\rvb} \!\in\! \sR^{d}$ (``$\leq$'' for vectors is elementwise here).
Note that for simplicity, we show a state feedback setting here; but we have also considered output feedback settings following \citet{yang2024lyapunov}, where a controller only has access to observed system outputs with state estimation by an observer, rather than the true state, and $\rvx_t$ is augmented to account for state estimation errors (discussed further in \Cref{ap:modeling}).

The goal is to guarantee that if the system starts from any state $\rvx\!\in\!\gS$ within some region-of-attraction (ROA), $\gS\!\subseteq\!\gB$, it will converge to the equilibrium $\rvx^*$.
To achieve this, a Lyapunov function $V(\rvx)\!:\!\sR^d\rightarrow\sR$ is to be learned jointly with the controller. The ROA is specified with a sublevel set within $\gB$, as $\gS\coloneqq \{ \rvx\in\gB \mid V(\rvx)<\rho\}$ with threshold $\rho$. We aim to have the following property verified:
\vspace{-10pt}
\begin{equation}
\forall\rvx_t\in\gB,
\enskip
V(\rvx_t)<\rho \rightarrow
\Big(V(\rvx_{t+1})-V(\rvx_t) \leq -\kappa V(\rvx_t)\Big) \wedge \big(\rvx_{t+1}\in\gB\big),
\label{eq:lyapunov}\vspace{-8pt}
\end{equation}
where $\kappa>0$ is a constant specifying the exponential stability convergence rate. When the positive definiteness of $V(\cdot)$, $V(\rvx^*)=0$ and $V(\rvx)>0~(\forall\rvx\neq\rvx^*)$, is guaranteed by construction as discussed in \Cref{ap:modeling}, and \eqref{eq:lyapunov} is verified for the entire ROI ($\forall\vx_t\!\in\!\gB$), the Lyapunov asymptotic stability is guaranteed for ROA $\gS$~\citep{yang2024lyapunov}. To clarify, since all possible states within $\gB$ are considered when verifying \eqref{eq:lyapunov}, the verification can be independent from time step $t$ and does not require reasoning over multi-step trajectories.

\vspace{-4pt}
\paragraph{Training problem.}
Both $u(\cdot)$ and $V(\cdot)$ are modeled as neural networks due to their expressive power, and we follow \citet{yang2024lyapunov} to use simple feedforward architectures as detailed in \Cref{ap:modeling}.
Suppose $g_\theta(\rvx_t)$ is a model that characterizes the violation of  \eqref{eq:lyapunov} on a given $\rvx_t$, parameterized by $\theta$. $g_\theta(\rvx_t)$ depends on $u(\cdot)$ and $V(\cdot)$, and state $\rvx_t$ is the input to the model.
Given \eqref{eq:lyapunov}, we have:
\vspace{-6pt}
\begin{equation}
g_\theta(\rvx_t)\coloneqq
\min~\bigg\{
\rho-V(\rvx_t),\,
\sigma(V(\rvx_{t+1})-(1-\kappa)V(\rvx_t))
+ \sum_{1\leq i\leq d}
\sigma([\rvx_{t+1}]_i-\ol{\rvb}_i)
+\sigma(\ul{\rvb_i}-[\rvx_{t+1}]_i)
\bigg\},
\label{eq:goal_lyap}\vspace{-8pt}
\end{equation}
where $\sigma(x)\coloneqq\max\{x,0\}$ is known as ReLU. We aim to achieve a verifiable
\vspace{-6pt}
\begin{equation}
    \forall \rvx_t\in\gB,
    \quad g_\theta(\rvx_t)\leq 0,
\label{eq:goal}\vspace{-6pt}
\end{equation}
which can be written as a min-max optimization problem as $\argmin_\theta\argmax_{\rvx_t\in\gB} g_\theta(\rvx_t)$, where $\theta$ denotes the parameters of both $u(\cdot)$ and $V(\cdot)$, jointly optimized during training. Note that \eqref{eq:goal} alone is insufficient: it could be trivially satisfied with a vacuously small ROA (e.g., by driving $V(\rvx)$ to be large everywhere so that $V(\rvx)<\rho$ holds for almost no state). An additional loss term to encourage a large ROA is therefore necessary, as discussed in \Cref{sec:model_lyap}.

\vspace{-4pt}
\subsection{CT-BaB: Certified Training Framework with Training-time Branch-and-bound}
\label{sec:training_framework}
\begin{figure}[t]
\centering
\includegraphics[width=\textwidth]{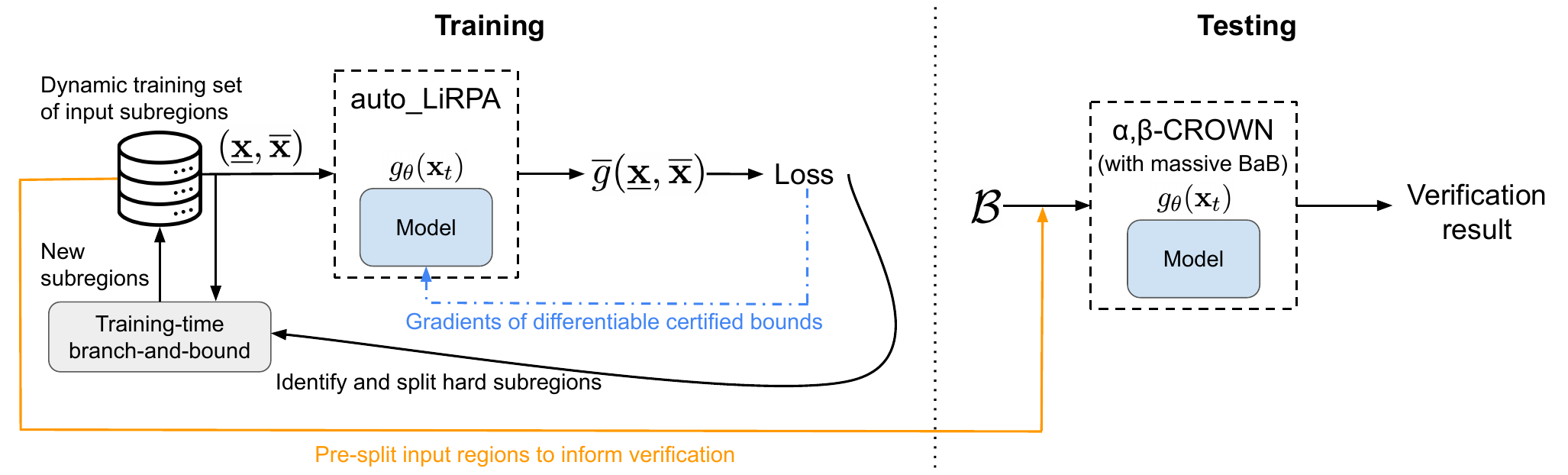}
\caption{Illustration of our proposed CT-BaB framework.}
\label{fig:framework}
\end{figure}

We illustrate our proposed CT-BaB framework in \Cref{fig:framework} and explain it below.

\paragraph{Certified upper bound for verification.}
We first explain the verification for \eqref{eq:goal}. Recent computationally efficient methods typically compute a certified bound $\ol{g}_\theta(\gB) $ such that $ \ol{g}_\theta(\gB) \!\geq\! g_\theta(\rvx_t)~(\forall\rvx_t\!\in\!\gB)$ provably holds, and then $\ol{g}(\gB)\leq 0$ implies a successful verification. State-of-the-art NN verifiers such as \abcrown\citep{zhang2018efficient,xu2020automatic,xu2020fast,wang2021beta,zhang2022general,shi2024genbab} leverage a technique called \emph{linear bound propagation}~\citep{zhang2018efficient,wong2018provable,singh2019abstract,xu2020automatic} for computing certified bounds, which relaxes nonlinear operators in the model by linear lower and upper bounds, and it then propagates linear bounds through the model to finally bound the output w.r.t. the input region. This technique has been generalized to support general computational graphs (i.e., general model architectures and problem specifications, including the problem here) in the auto\_LiRPA framework~\citep{xu2020automatic} used by \abcr. Additionally, since initial certified bounds may not be tight enough for a successful verification, due to the over-approximation from linear relaxation, a branch-and-bound (BaB) technique is also commonly used by branching the verification problem into smaller subproblems and computing tighter certified bounds for smaller subproblems~\citep{bunel2018unified,bunel2020branch,wang2021beta}. Since current works on this Lyapunov stability task commonly focus on systems with low-dimensional inputs, \citet{yang2024lyapunov} configured \abcrown to run BaB by splitting the state (i.e., input $\rvx_t$ to model $g_\theta(\rvx_t)$, not internal activations)\footnote{Although \abcrown usually conducts BaB by splitting activations and did not explicitly discuss the use of input splitting in papers, input splitting has in fact been used to verify many low-dimensional systems.}.

\vspace{-6pt}
\paragraph{Limitation of CEGIS.}
In previous works using CEGIS~\citep{wu2023neural,yang2024lyapunov}, each training iteration basically searches for counterexamples $\rvx^{(1)}_t, \rvx^{(2)}_t, \cdots\in\gB$ such that $g_\theta(\rvx^{(1)}_t)$, $g_\theta(\rvx^{(2)}_t),\cdots>0$. Then it optimizes the model on a finite number of counterexamples as $ \argmin_\theta$ $ \big(g_\theta(\rvx^{(1)}_t)+g_\theta(\rvx^{(2)}_t)+\cdots\big)$. While it often eliminates noticeable counterexamples during training, i.e., achieving $ g_\theta(\rvx)\leq 0$ for concrete data points $\rvx$, they generally do not make certified bounds $ \ol{g}(\gB)$ tight enough for efficient verification. This is because the computation of certified bounds is not considered in training, and there is a significant discrepancy between $g_\theta(\cdot)$ optimized during training and certified bounds $\ol{g}_\theta(\cdot)$ computed at test time. This limitation leads to extensive BaB required at test time to tighten certified bounds and thus costly verification, particularly given the difficulty of verification for NNs~\citep{salman2019convex}.

\vspace{-6pt}
\paragraph{Certified training for verification-friendly models.} In contrast, we propose to conduct certified training, where we incorporate the computation of verification into the training and optimize certified bounds. Specifically, we essentially optimize the model by $\argmin_\theta \sigma(\ol{g}_\theta(\gB))$, where $\ol{g}_\theta(\gB)$ is computed by  auto\_LiRPA~\citep{xu2020automatic} with linear bound propagation, thereby reducing the discrepancy between training and test-time verification and making trained models more verification-friendly. This training is possible because $ \ol{g}_\theta(\gB)$ computed by linear bound propagation is differentiable, as both linear relaxation and each linear bound propagation step consist of differentiable operations that can be used in  training~\citep{zhang2019towards,xu2020automatic}.

\vspace{-6pt}
\paragraph{Training-time branch-and-bound.}
Previous certified training works~\citep{gowal2018effectiveness,mirman2018differentiable,shi2021fast,wang2021convergence,muller2022certified,de2022ibp,mao2024connecting} commonly focused on adversarial robustness within small local neighborhoods, e.g., $ \{\rvx: \|\rvx-\rvx_0\| \!\leq\!\eps\}$ around each individual example $\rvx_0\!\in\!\gB$ (suppose $\gB$ is still the set of all possible inputs), with a small radius $\eps$~\citep{goodfellow2014explaining}. For them, each certified bound computation only needs to handle a small local region within $\gB$. However, unlike those previous works, we desire guarantees on the entire $\gB$ (as \eqref{eq:goal}). Since $\gB$ covers all the states that the model cares about, it is large and relatively global, and directly computing $\ol{g}_\theta(\gB)$ leads to extremely loose certified bounds. Naively, one may split $\gB$ into many small subregions in the beginning. However, this would be inefficient, as the difficulty of computing relatively tight certified bounds varies for different subregions, and uniformly splitting $\gB$ in the beginning cannot identify and spend much more splitting on the hardest subregions. To address this, we propose a novel \emph{training-time branch-and-bound} technique. Throughout the training, we dynamically split $\gB$ into smaller subregions, and we maintain a dataset with $n$ examples:
$\sD=\{
(\ul{\rvx}^{(1)}, \ol{\rvx}^{(1)}),
(\ul{\rvx}^{(2)}, \ol{\rvx}^{(2)}),
\cdots,
(\ul{\rvx}^{(n)}, \ol{\rvx}^{(n)})
\},$
where each example $(\ul{\rvx}^{(k)}, \ol{\rvx}^{(k)})~(1\leq k\leq n)$ is a subregion in $\gB$, defined as a bounding box $\{\rvx: \rvx\!\in\!\sR^{d},\,\ul{\rvx}^{(k)} \!\leq\! \rvx \!\leq\! \ol{\rvx}^{(k)}\}$.
All these examples are non-overlapping and cover $\gB$.
We dynamically update and expand the dataset during training, by splitting hard examples into more examples with even smaller subregions, as detailed in \Cref{sec:bab}.

\vspace{-6pt}
\paragraph{Empirical training objective.}
In addition to the main training objective with certified bounds, we also add a term where we try to empirically find the worst-case violation of \eqref{eq:goal} by projected gradient descent (PGD)~\citep{madry2017towards}. It can make the training more quickly reach a solution with most counterexamples eliminated, for the certified training to focus on making it verifiable. It also helps to achieve that at least no counterexample can be empirically found, even when not all the subregions in $\sD$ can be verified yet, as we may fully verify \eqref{eq:goal} using \abcrown with more extensive BaB at test time.
Specifically, we denote this objective as $ \ol{g}^{A}(\ul{\rvx},\ol{\rvx})\coloneqq g(A(\ul{\rvx},\ol{\rvx}))$, where $ A(\ul{\rvx},\ol{\rvx})~(\ul{\rvx} \leq A(\ul{\rvx},\ol{\rvx}) \leq \ol{\rvx})$ is a data point empirically found by PGD to maximize $g(A(\ul{\rvx},\ol{\rvx}))$,
as $\argmax_{\rvx\in\sR^d~(\ul{\rvx} \leq \rvx\leq \ol{\rvx})} g(\rvx)$, but this maximization by PGD is not guaranteed to be optimal.

\vspace{-2pt}
\paragraph{Overall training objective.}
Overall, we optimize for the following loss function:
\vspace{-7pt}
\begin{equation}
L =
\frac{1}{|\sD|}\sum_{(\ul{\rvx},\ol{\rvx})\in\sD}\enskip
\Big(\sigma(\ol{g}(\ul{\rvx},\ol{\rvx}) + \eps)
+
\lambda\sigma(\ol{g}^{A}(\ul{\rvx},\ol{\rvx}) + \eps)
\Big)
+ L_{\text{extra}},
\quad\text{where}\
\bigcup_{(\ul{\rvx},\ol{\rvx})\in\sD}\!=\!\gB,
\label{eq:loss}\vspace{-7pt}
\end{equation}
$\eps$ is small value for ideally eliminating the violation with a margin, $\lambda$ is a coefficient for weighting the PGD term, and $L_{\text{extra}}$ is an extra loss term for controlling additional properties as discussed in \Cref{sec:model_lyap}.
We monitor training convergence by tracking the proportion of verified subregions in $\sD$ (i.e., those with $\ol{g}(\ul{\rvx}^{(k)}, \ol{\rvx}^{(k)}) \leq 0$).
After the training, we use \abcrown to verify \eqref{eq:goal}, and thus the soundness of the trained models is guaranteed as long as the final verification succeeds.
It is worth noting that CT-BaB is generally formulated, as $g(\cdot)$ can potentially model other properties from various applications, but we focus on Lyapunov asymptotic stability in this work.

\vspace{-6pt}
\subsection{Training-Time Branch-and-Bound}
\label{sec:bab}

As discussed in \Cref{sec:training_framework}, our training-time BaB aims to address challenges in obtaining relatively tight certified bounds when we require guarantees on the entire $\gB$. We introduce it in more detail.

\vspace{-4pt}
\paragraph{Initial splits.}

We initialize $\sD$ by splitting $\gB$ into grids along each of its $d$ dimensions, respectively. We control the maximum size of the initial regions with a threshold $l$ that denotes the maximum length of each input dimension. For each input dimension $i~(1\!\leq\! i\!\leq\! d)$, we uniformly split the input range $[\ul{\rvb}_i,\ol{\rvb}_i]$ into $m_i= \lceil \frac{\ol{\rvb}_i-\ul{\rvb}_i}{l} \rceil$ segments in the initial split, such that the length of each segment is no larger than the threshold $l$.
We thereby create $ \prod_{i=1}^d m_i$ regions to initialize $\sD$.
We set $l$ such that the number of initial examples is close to the batch size.

\vspace{-5pt}
\paragraph{Dynamic splits in training.}
During training, we dynamically split hard regions into even smaller subregions after every epoch. For each training batch, we take each $(\ul{\rvx}^{(k)}, \ol{\rvx}^{(k)})$ with $ \ol{g} (\ul{\rvx}^{(k)}, \ol{\rvx}^{(k)}) \!>\! 0 $, i.e., we have not been able to verify that $g(\rvx)\!\leq\! 0$ for region $[\ul{\rvx}^{(k)}, \ol{\rvx}^{(k)}]$. We then uniformly split the region into two subregions along a chosen input dimension $i(1\leq i\leq d)$,
and we replace the original region with the two new subregions in the dataset.
To maximize the benefit of splitting an example, we choose the input dimension by considering each input dimension $j(1\leq j\leq d)$ and computing the total loss of the two new subregions when dimension $j$ is split.
Suppose $L(\ul{\rvx}^{(k)},\ol{\rvx}^{(k)})$ is the contribution of an example $(\ul{\rvx}^{(k)},\ol{\rvx}^{(k)})$ to the loss function in \eqref{eq:loss}.
We take the dimension $j$ that leads to the lowest loss value for the new examples to split:
\vspace{-15pt}
\begin{equation}
\argmin_{1\leq j\leq d} \enskip L(\ul{\rvx}^{(k)}, \ol{\rvx}^{(k,j)})  + L(\ul{\rvx}^{(k,j)}, \ol{\rvx}^{(k)}),
\quad \text{where}\enskip
 \ul{\rvx}^{(k,j)}_j=\ol{\rvx}^{(k,j)}_j= \frac{\ul{\rvx}^{(k)}_j+\ol{\rvx}^{(k)}_j}{2},
\label{eq:split}
\vspace{-6pt}
\end{equation}
and $\ul{\rvx}^{(k,j)}_i \!=\!\ul{\rvx}^{(k)}_i,  \ol{\rvx}^{(k,j)}_i \!=\!  \ol{\rvx}^{(k)}_i$ keep unchanged for other dimensions $i\!\neq\! j$. This only mildly affects the total training cost. As training progresses, most  subregions are verified and thus no longer require splitting, and different regions to split and input dimensions to consider can be handled in parallel on GPUs, though further training may introduce new counterexamples in previously verified subregions, requiring those subregions to be split again.
The splitting may terminate primarily when the dataset $\sD$ reaches a maximum size, as the system memory can be limited (in our experiments, the limit is never reached). Optionally, splitting can also be stopped after a fixed number of training steps, which is not required in our experiments.

\vspace{-4pt}
\paragraph{Training-aware verification.}
As mentioned in \Cref{sec:training_framework}, \abcrown with more extensive BaB is used at test time to finally verify the model. When the training converges, CT-BaB can leave a small proportion of training subregions unverified (e.g., up to 6\% in our results), and these subregions require further splitting.
By default, \abcrown starts splitting from the original $\gB$ and is likely to take different splitting paths compared to our training-time BaB. To further reduce the discrepancy between training and testing, we propose \emph{training-aware verification} by exporting training subregions to inform test-time verification. We modify the \abcrown verifier to allow loading so-called ``pre-split'' input regions, i.e., input regions from our final $\sD$ that have already been split from $\gB$ during training prior to verification. Thereby, our training-time BaB can bootstrap and further speed-up the test-time verification.

\subsection{Controlling the ROA}
\label{sec:model_lyap}

As shown in \eqref{eq:loss}, an additional term $L_{\text{extra}}$ can be added to control additional properties. We use this term to control the size of the region-of-attraction (ROA). We aim to have a good proportion of data points $\rvx\in\gB$ that are within the ROA as $V(\rvx)<\rho$. We randomly draw $n_\rho$ samples within $\gB$, as $\tilde{\rvx}_1,\tilde{\rvx}_2,\cdots,\tilde{\rvx}_{n_\rho}\in\gB$ (these samples are independent from $(\ul{\rvx}^{(k)}, \ol{\rvx}^{(k)})$ used for other terms in \eqref{eq:loss}), and we define $L_{\text{extra}}$ as:
\vspace{-8pt}
\begin{equation}
L_{\text{extra}} =
\sI\bigg(\frac{1}{n_\rho}\sum_{i=1}^{n_\rho}\sI(V(\tilde{\rvx}_i)<\rho)<\rho_{\text{ratio}}\bigg)
\frac{\lambda_\rho}{n_\rho}\sum_{i=1}^{n_\rho} \sigma(V(\tilde{\rvx}_i)+\rho-\eps),
\label{eq:roa}\vspace{-6pt}
\end{equation}
which penalizes samples with $V(\tilde{\rvx}_i)>\rho-\eps$ when the ratio of samples within the sublevel set is below the threshold $\rho_{\text{ratio}}$, thereby encouraging more samples to be within the ROA. Here $\eps$ is a small value for the margin as similarly used in \eqref{eq:loss} and $\lambda_\rho$ is for weighting the $L_{\text{extra}}$ term in \eqref{eq:loss}. In our implementation, we simply fix $\rho=1$ and make $n_\rho$ equal to the batch size. The threshold $\rho_{\text{ratio}}$ and the weight $\lambda_{\rho}$ can be set to reach the desired ROA size, although setting a stricter requirement on ROA naturally tends to increase the difficulty of training. Compared to the loss term for ROA in \citet{yang2024lyapunov} which required carefully selecting candidate states that are desired to be within the ROA by referring to classical LQR solutions, ours is self-contained and does not refer to any other solution, and thus can reduce the burden of applying our method. Note that this term is a necessary component of the training: without it, the Lyapunov condition \eqref{eq:lyapunov} could be trivially satisfied by collapsing the ROA to a negligibly small region.

\section{Experiments}

Additional experimental details are included in \Cref{ap:details}.

\begin{table}[ht]
\centering
\caption{Dynamical systems in the experiments, all following previous works~\citep{yang2024lyapunov,wu2023neural}.
$d$ and $n_u$ represent the dimension of input state and control input, respectively.
There is a limit on $u$ which is clamped according to the limit, where some symbols are from the system dynamics: $m$ for mass, $g$ for gravity, $l$ for length, and $v$ for velocity.
Size of the region-of-interest is denoted by the upper boundary $\ol{\rvb}$ (with $\ul{\rvb}=-\ol{\rvb}$).
Equilibrium state of all the systems is $\rvx^*=\vzero$.
Systems with ``(output)'' are output feedback settings, and others are state feedback settings.
}
\adjustbox{max width=.98\textwidth}{
\begin{tabular}{lccccccc}
\toprule
System & $d$ & $n_u$ & Limit on $u$ & \makecell{Region-of-interest\\(denoted as upper bound of box)} \\
\midrule
\multirow{2}{*}{Inverted Pendulum} & \multirow{2}{*}{2} & \multirow{2}{*}{1} & $|u|\leq 8.15\cdot mgl$ (large torque) & \multirow{2}{*}{$[12, 12]$} \\
& & & $ |u|\leq 1.02\cdot mgl$ (small torque)\\
Inverted Pendulum (output) & 4 & 1 & $|u|\leq \frac{mgl}{3}$ & $[0.7\pi, 0.7\pi, 0.175\pi, 0.175\pi]$\\
\multirow{2}{*}{Path Tracking} & \multirow{2}{*}{2} & \multirow{2}{*}{1} & $ |u| \leq 1.68\cdot l/v$ (large torque) & \multirow{2}{*}{$[3,3]$}  \\
& & & $ |u|\leq l/v$ (small torque)\\
Cart-Pole & 4 & 1 & $|u|\leq 30$ & $[1, 1, 1, 1]$\\
PVTOL & 6 & 2 & $\|u\|_\infty \leq mg$ & $[0.75, 0.75, \pi,4,4,3]$ \\
2D Quadrotor & 6 & 2 & $\|u\|_\infty\leq 1.25\cdot mg$  & $[0.75, 0.75, \pi,4,4,3]$\\
2D Quadrotor (output) & 8 & 2 & $\|u\|_\infty\leq 1.5\cdot mg$ & $[0.1,0.2\pi,0.2,0.2\pi,0.05,0.1\pi,0.1,0.1\pi]$\\
\bottomrule
\end{tabular}}
\label{tab:systems}
\end{table}

\vspace{-10pt}
\paragraph{Dynamical systems.}
We experiment on several dynamical systems following \citet{wu2023neural,yang2024lyapunov}, as listed in \Cref{tab:systems}:
\emph{Inverted Pendulum} is about swinging up a pendulum to the upright equilibrium;
\emph{Path Tracking} is about tracking the path for a planar vehicle;
\emph{Cart-Pole} is about balancing an Inverted Pendulum mounted on a Cart-Pole;
\emph{PVTOL} is about drone planar vertical takeoff and landing;
and \emph{2D Quadrotor} is about hovering a planar quadrotor.
For Inverted Pendulum and Path Tracking, there are two different limits on the maximum allowed torque for the controller, where the setting is more challenging with a smaller torque limit.
Unless otherwise noted, state feedback is considered for all these systems. Output feedback is additionally considered for Inverted Pendulum and 2D Quadrotor, following \citet{yang2024lyapunov}. Their system dynamics are attached at \Cref{ap:system_dynamics}.

\vspace{-6pt}
\paragraph{Baselines and Metrics.}
We compare with previous works on the same task, i.e., learning NN-based controllers with verified Lyapunov asymptotic stability for discrete-time dynamical systems~\citep{wu2023neural,yang2024lyapunov}. Among them, \citet{yang2024lyapunov} is the previous state-of-the-art, and they both outperform classical non-learning methods such as LQR~\citep{tedrake2010lqr}. For comparison, we mainly use the verification time cost and size of ROA as the metrics. We follow \citet{wu2023neural} to estimate the size of ROA. We consider grid points in the region-of-interest $\gB$ and count the proportion of grid points within the verified $\gS$, multiplied by the volume of $\gB$. Hyperparameters for all the model architectures follow \citet{yang2024lyapunov}.

\begin{table}[t]
\centering
\caption{
Comparison on the verification time cost and the size of ROA. Model checkpoints for \citet{wu2023neural} are obtained from the source code of \citet{yang2024lyapunov} and the same models have been used for comparison in \citet{yang2024lyapunov}. ``-'' denotes that the models for \citet{wu2023neural,yang2024lyapunov} are not available on some of the systems (although \citet{wu2023neural} originally had models for PVTOL, an issue was found by \citet{yang2024lyapunov} and the training could not work after the issue was fixed).
``+Loading'' indicates that we load training subregions into \abcrown for a training-aware verification, as discussed in \Cref{sec:bab}.
}
\adjustbox{max width=.99\textwidth}{
\begin{tabular}{lccccccc}
\toprule
\multirow{2}{*}{System} &
\multicolumn{2}{c}{{\citet{wu2023neural}}}
&
\multicolumn{2}{c}{CEGIS~\citep{yang2024lyapunov}} &
\multicolumn{3}{c}{CT-BaB (Ours)} \\
& Time & ROA & Time & ROA & \multicolumn{2}{c}{Time} & ROA\\
& & & & & & (+Loading) & \\
\midrule
Inverted Pendulum (large torque) & {11.3s} & {53.28} & 33s & 239.04 & 12.4s & \textbf{1.8s} & \textbf{505.4} \\
Inverted Pendulum (small torque) & - & - & 25s &  187.20 & 12.0s & \textbf{2.2s} & \textbf{457.9}\\
Inverted Pendulum (output) & - & - & 94s & 6.26 & 48.8s & \textbf{5.3s} & \textbf{12.23} \\
Path Tracking (large torque) & {11.7s} & 14.38 & 39s & 18.27 & 19.2s & \textbf{3.0s} & \textbf{20.79}\\
Path Tracking (small torque) & - & - & 34s & 10.53 & 17.2s & \textbf{3.6s} & \textbf{14.03}\\
Cart-Pole & 2.2min & 0.037 & - & - & 68s & \textbf{2.7s} & \textbf{1.35}\\
PVTOL & - & - & - & - & 109s & \textbf{16s} & \textbf{46.95}\\
2D Quadrotor & - & - & 1.1hrs & 3.29 & 144s & \textbf{49s} & \textbf{54.39} \\
2D Quadrotor (output) & - & - & 8.9hrs & 6.7e-9 & $>$12hrs & \textbf{0.8hrs} &  \textbf{1.1e-6}\\
\bottomrule
\end{tabular}
}
\label{tab:mainresults}
\end{table}

\vspace{-6pt}
\paragraph{Main Results.} We show the main results in \Cref{tab:mainresults}. On all the settings where any baseline is applicable, CT-BaB yields much larger ROA and much smaller verification time cost. Among the experimented systems, Inverted Pendulum, Path Tracking, and Cart-Pole are relatively small in terms of the number of input states (up to 4), where test-time verification for our models completes within 4s. For PVTOL, previously only \citet{wu2023neural} trained a model but their training could not work after an implementation issue was found and fixed by \citet{yang2024lyapunov}; our training works successfully on PVTOL. On the two largest systems, 2D Quadrotor with state feedback and output feedback, respectively, our improvement over the previous state-of-the-art \citet{yang2024lyapunov} is particularly significant. Specifically, for the state feedback setting, we reduce the verification time from 1.1hrs to 49s, while enlarging the ROA by $16\times$; for the output feedback setting, we reduce the verification time from 8.9hrs to 0.8hrs, while enlarging the ROA by $164\times$. 
We provide a visualization of ROA in \Cref{ap:roa}.
These results demonstrate the significant advantage of our CT-BaB method for producing verification-friendly models (i.e., models that can be verified in much shorter time) with stronger verifiable guarantees (i.e., larger ROA where the Lyapunov stability is guaranteed).

\vspace{-4pt}
\paragraph{Advantage of training-aware verification.}
As proposed in \Cref{sec:bab}, our training-aware verification is informed by training-time BaB and loads training subregions. In \Cref{tab:mainresults}, the results show that loading training subregions significantly reduces the verification time cost on all the settings. 
On the largest 2D Quadrotor with output feedback, the original verification times out after trying verification for 12 hours, which is reasonable considering that our ROA is $164\times$ larger than the one in \citet{yang2024lyapunov} verified with 8.9 hours; in contrast, by loading our training subregions, the verification completes within 0.8 hours. 
On other systems, even without loading training subregions, our verification time cost is still much smaller compared to \citet{yang2024lyapunov} while our ROA is much larger, which similarly holds for Cart-Pole where only results from \citet{wu2023neural} are available. 
Although \citet{wu2023neural} only took approximately 11s to verify systems on Inverted Pendulum and Path Tracking (large torque), their ROA is much smaller compared to \citet{yang2024lyapunov} and ours. Overall, our results demonstrate that CT-BaB not only produces verification-friendly models but also enables training-aware verification to further speed up the verification.

\begin{table}[ht]
\centering
\caption{
Runtime of training, the size of the training dataset, the ratio of training subregions verifiable without further BaB at the end of the training, and the number of BaB domains visited by \abcrown during verification. All the models are fully verified by \abcrown with more BaB.
}
\adjustbox{max width=.98\textwidth}{
\begin{tabular}{lcccccc}
\toprule
\multirow{2}{*}{System} & \multirow{2}{*}{Runtime} & \multicolumn{2}{c}{Dataset size} & \multicolumn{3}{c}{Verified} \\
& & Initial & Final & W/o more BaB & By \abcrown & BaB domains\\
\midrule
Inverted Pendulum (large torque) & 1.7min & 58,080 & 68,686 & 99.9993\% & 100\% & 12\\
Inverted Pendulum (small torque) & 11.0min & 58,080 & 2,286,676 & 99.91\% & 100\% & 14,196\\
Inverted Pendulum (output) & 11.8min & 19,200 & 3,411,802 & 99.7\% & 100\% & 134,187\\
Path Tracking (large torque) & 3.3min & 40,400 & 596,100 & 98.4\% & 100\% & 60,079\\
Path Tracking (small torque) & 14.9min & 40,400  & 1,325,095 & 99.6\% & 100\% & 78,144\\
Cart-Pole & 48.7min & 12,480 & 3,795,905 & 99.95\% & 100\% & 17,940\\
PVTOL & 1hrs & 46,336 & 18,977,973 & 95.6\% & 100\% & 40,603,447\\
2D Quadrotor & 1.5hrs & 46,336 & 29,015,573 & 95.6\% & 100\% & 154,390,951\\
2D Quadrotor (output) & 2.0hrs & 44,032 & 20,448,758 & 94.0\% & 100\% & 2,745,392,253\\
\bottomrule
\end{tabular}}
\label{tab:trainingdetails}
\end{table}

\vspace{-4pt}
\paragraph{Training time and data.}
In \Cref{tab:trainingdetails}, we show more information about the training, including the time cost of training and size of the dynamic training dataset.
Our training dataset is dynamically maintained as described in \Cref{sec:bab}, and the dataset size grows from the ``initial'' size to the ``final'' size. At the end of the training, most training subregions (at least 94.0\%) can be verified without further BaB, and all the models can be fully verified by \abcrown with more extensive BaB at test time.
A detailed comparison of runtime against \citet{yang2024lyapunov} is provided in \Cref{tab:trainingtime}; although the training time of CT-BaB is often higher than \citet{yang2024lyapunov}, CT-BaB achieves the lowest total time cost on the largest 2D Quadrotor with output feedback setting, demonstrating its potential for scaling up.

\section{Related Work}
\label{sec:related}

\paragraph{Lyapunov-stable controllers.}
Some works on Lyapunov stability focused constructing Lyapunov functions for a given dynamical system, without jointly learning a controller~\citep{alfarano2023discovering,zou2025analytical,liu2025physics,giesl2026kernel}.
Classical methods such as linear quadratic regulator (LQR)~\citep{tedrake2010lqr} and sum-of-squares (SOS)~\citep{parrilo2000structured,majumdar2013control,yang2023approximate,dai2023convex} can synthesize stable controllers with guarantees but are oftentimes limited to linear or polynomial controllers. In contrast, learning NN-based controllers has shown great potential for more complicated systems and larger ROA. Many works only used sampling without formal guarantees~\citep{jin2020neural,sun2021learning,dawson2022safe,liu2023safe,min2023data}. Although some works such as \citet{jin2020neural} theoretically considered verification, they assumed the existence of a Lipschitz constant that was not practically computed.

To achieve formal verification for NN-based controllers, previous works commonly adopted a Counter-Example Guided Inductive Synthesis (CEGIS) approach by iteratively searching for counterexamples and optimizing the models to eliminate the counterexamples. The counterexamples can be generated by Satisfiable Modulo Theories (SMT) solver~\citep{gao2013dreal,de2008z3,chang2019neural,abate2020formal}, Mixed Integer Programming (MILP) solver~\citep{dai2021lyapunov, chen2021learning,wu2023neural}, or more efficiently, projected gradient descent (PGD)~\citep{madry2017towards,wu2023neural,yang2024lyapunov}. Among them, \citet{wu2023neural} used a verifier~\citep{xu2020automatic} but only to guarantee the positive definiteness of the Lyapunov function (which can also be directly achieved by construction, as done in \citet{yang2024lyapunov}) and not the main Lyapunov condition.
To our knowledge, we are the first to incorporate certified bounds for the violation of the Lyapunov condition during training, to produce verification-friendly models with stronger guarantees, while informing test-time verification for a training-aware verification.

\paragraph{Safety verification beyond stability.} Apart from Lyapunov stability, there are also many works on verifying other safety properties of neural controllers, such as reachability~\citep{althoff2016cora,tran2020nnv,hu2020reach,everett2021reachability,ivanov2021verisig,knuth2021planning,huang2022polar,wang2023polar,schilling2022verification,chatterjee2023learner,kochdumper2023open,jafarpour2023interval,jafarpour2024efficient,teuber2024provably,badings2025policy}, forward invariance~\citep{ames2016control,zhao2021learning,wang2023enforcing,huang2023fi,harapanahalli2024certified,hu2024verification,wang2024simultaneous}, and contraction~\citep{tsukamoto2020neural,fitzsimmons2024computation,li2025neural}. The Lyapunov asymptotic stability we study is a relatively strong guarantee, as it ensures convergence towards an equilibrium.

\section{Conclusion}
\label{sec:conc}
To conclude, we propose CT-BaB for training Lyapunov asymptotically stable NN-based controllers and achieving verifiable guarantees on an entire input region-of-interest. CT-BaB is verification-aware by optimizing certified bounds during training, and it handles a relatively large region-of-interest by a new training-time BaB. It also informs test-time verification by the dynamic training dataset created during training, for a more efficient training-aware verification. We have demonstrated that CT-BaB consistently produces more verification-friendly models with stronger guarantees, i.e., much smaller verification cost at test time with much larger ROA.

A limitation of this work is that only low-dimensional dynamical systems have been considered, which is also a common limitation of previous works on Lyapunov stability~\citep{chang2019neural,wu2023neural,yang2024lyapunov}. Even after the training, the verification may still fail. In this case, like prior works, we may continue training, as further optimization of the certified bounds may yield a more verification-friendly model. However, there may be fundamental challenges that prevent successful verification, for instance when the system dimensionality is too large. Future works may try to scale up our method to higher-dimensional systems, by potentially conducting training-time BaB on learned internal activations, not only inputs. Additionally, although we focus on Lyapunov stability in this work, our CT-BaB is generally formulated, and thus we envision that CT-BaB can potentially be applied to other types of safety properties in future work.

\section*{Acknowledgments}
This project is supported in part by NSF 2048280, 2331966 and ONR N00014-23-1-2300:P00001. 
Huan Zhang is supported in part by the AI2050 program at Schmidt Sciences (AI2050 Early Career Fellowship) and NSF (IIS-2331967). 

\bibliography{papers}

\newpage
\appendix
\section{Details of the Implementation and Experiments}
\label{ap:details}

\subsection{Modeling}
\label{ap:modeling}

Since our focus is on a new training framework, we follow model architectures in \citet{yang2024lyapunov}, and we follow their source code which has some minor difference with details mentioned in their paper.

\paragraph{Controller.}
We use a fully-connected NN for the controller $u(\rvx)$. There are 8 hidden neurons in each hidden layer. For Inverted Pendulum and Path tracking, there are 4 layers, and other systems, there are 2 layers. ReLU is used as the activation function.

\paragraph{Lyapunov function.}
For the Lyapunov function $V(\rvx)$, we either use a model based on a fully-connected NN $\phi(\rvx)$, as $ V(\rvx)=|\phi(\rvx)-\phi(\rvx^*)|+\|(\eps_V \rmI+\rmR^\top \rmR)(\rvx-\rvx^*)\|_1$, or a quadratic function as $ V(\rvx)=(\rvx-\rvx^*)^\top (\eps_V \rmI +\rmR^\top \rmR)(\rvx-\rvx^*)$, where $\rmR\in\sR^{n_r\times n_r}$ is an optimizable matrix parameter, and $\eps_V>0$ is a small value to guarantee positive definiteness. This construction automatically guarantees $V(\rvx^*)=0$ and $V(\rvx)>0~(\forall \rvx\neq\rvx^*)$~\citep{yang2024lyapunov}. A NN-based Lyapunov function is used for Inverted Pendulum and Path tracking, where the NN is a fully-connected NN with 4 layers, and the number of hidden neurons is 16, 16, and 8 for the three hidden layers, respectively. Leaky ReLU is used as the activation function for NN-based Lyapunov functions. A quadratic Lyapunov function with $n_r=d$ is used for other systems.
Our framework is general w.r.t. the formulation of the Lyapunov function. Regardless of the formulation of $V(\rvx)$, it can be implemented as a module in PyTorch with the corresponding parameters, and jointly trained with the controller by optimizing \eqref{eq:loss}, where the Lyapunov function is simply part of the PyTorch computational graph specified by \eqref{eq:loss}.

\paragraph{Output feedback system and observer.}
There are two different settings considered -- state feedback control and output feedback control. The state feedback setting is straightforward, and the true state is directly observed as $\rvx_t$. In the output feedback setting, the controller is accompanied by an NN-based observer taking the observable output of the system $\rvy_t$ and predicts an estimated state $\hat{\rvx}_t$, and the controller then only takes the estimated state $\hat{\rvx}_t$ and system output $y_t$ rather than the true state $\tilde{\rvx}_t$; and the input state $\rvx_t\!=\![\tilde{\rvx}_t, \rve_t]$ of the problem models the true state $\tilde{\rvx}_t$ and the state estimation error $\rve_t\!=\!\hat{\rvx}_t\!-\!\tilde{\rvx}_t$. Thus, output feedback settings contain more input dimensions and an additional observer NN than their state feedback counterparts. The observer is a fully-connected NN with 3 layers for Inverted Pendulum and 2 layers for 2D Quadrotor, with 8 neurons in each hidden layer. System dynamics for output feedback systems are included in \Cref{ap:dynamics_output}, and we refer readers to Section 2 and Section 4.3 in \citet{yang2024lyapunov} for further details.

\subsection{Training}
We use a batch size of 30000 for all the training, with a learning rate of $5\times 10^{-3}$. We train models for up to 15,000 steps, but training for relatively smaller systems converges much faster. In the loss function, we set $\lambda$ to $10^{-4}$, $\lambda_p$ to 0.1, and $\eps$ to $10^{-4}$.
The only hyperparameter we tune for individual systems is $\rho_{\text{ratio}}$: we try to make it as large as possible for individual systems to maximize ROA, as long as the training works, which is principled.
For Inverted Pendulum and Path tracking, the range of $\rho_{\text{ratio}}$ is between 0.6 and 0.9; for other systems, we set $\rho_{\text{ratio}}=0.1$.
We start our dynamic splits after 100 initial training steps. We use auto\_LiRPA~\citep{xu2020automatic} to compute certified bounds, and we configure it to mainly use the backward bound propagation algorithm (a.k.a., CROWN~\citep{zhang2018efficient} generalized to general computational graphs), while we use Interval Bound Propagation (IBP)~\citep{gowal2018effectiveness,mirman2018differentiable} for bounding hidden layers of NNs required for the linear relaxation in CROWN, following \citet{zhang2019towards,xu2020automatic}. For PGD, we use 10 steps and a step size of 0.25 relative to the size of the subregion.

\section{Additional Results}
\label{ap:additional_results}

\subsection{Detailed Comparison of Runtime}

\begin{table}[ht]
\centering
\caption{
Comparison of training time, verification time, and total time between CT-BaB (ours) and CEGIS~\citep{yang2024lyapunov}.
Training times for \citet{yang2024lyapunov} are obtained by running their released code and are only available for the listed settings.
}
\adjustbox{max width=.99\textwidth}{
\begin{tabular}{lcccccc}
\toprule
\multirow{2}{*}{System} &
\multicolumn{3}{c}{CEGIS~\citep{yang2024lyapunov}} &
\multicolumn{3}{c}{CT-BaB (Ours)} \\
& Training & Verification & Total & Training & Verification & Total \\
\midrule
Inverted Pendulum           & 8.4min          & 33s    & 8.9min          & \textbf{1.7min} & \textbf{1.8s}   & \textbf{1.7min} \\
Inverted Pendulum (output)  & \textbf{7.3min} & 94s    & \textbf{8.8min} & 11.8min         & \textbf{5.3s}   & 11.9min \\
Path Tracking               & 12.4min         & 39s    & 13.1min         & \textbf{3.3min} & \textbf{3.0s}   & \textbf{3.4min} \\
2D Quadrotor                & \textbf{8.1min} & 1.1hrs & \textbf{1.24hrs}& 1.5hrs          & \textbf{49s}    & 1.51hrs \\
2D Quadrotor (output)       & \textbf{7.5min} & 8.9hrs & 9.02hrs         & 2.0hrs          & \textbf{0.8hrs} & \textbf{2.8hrs} \\
\bottomrule
\end{tabular}}
\label{tab:trainingtime}
\end{table}

In \Cref{tab:trainingtime}, we provide a detailed runtime comparison with \citet{yang2024lyapunov}.
Although our verification time is consistently lower than \citet{yang2024lyapunov}, they sometimes achieve lower training time and total time. However, on the largest system of 2D Quadrotor with output feedback, our total time is also much less than \citet{yang2024lyapunov} due to a large improvement on verification time. This demonstrates the potential of our work for further scaling up to larger systems.

\subsection{Visualization of ROA}
\label{ap:roa}
\begin{figure}[ht]
\centering
\includegraphics[width=.35\textwidth]{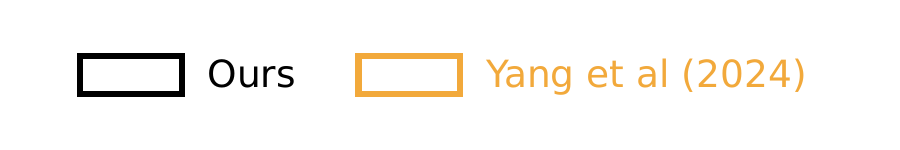}\\
    \vspace{-5pt}
    \subfigure[Projected to $y$ and $\theta$.]{%
        \begin{tikzpicture}
            \node[anchor=south west,inner sep=0] (image) at (0,0) {\includegraphics[width=0.29\textwidth]{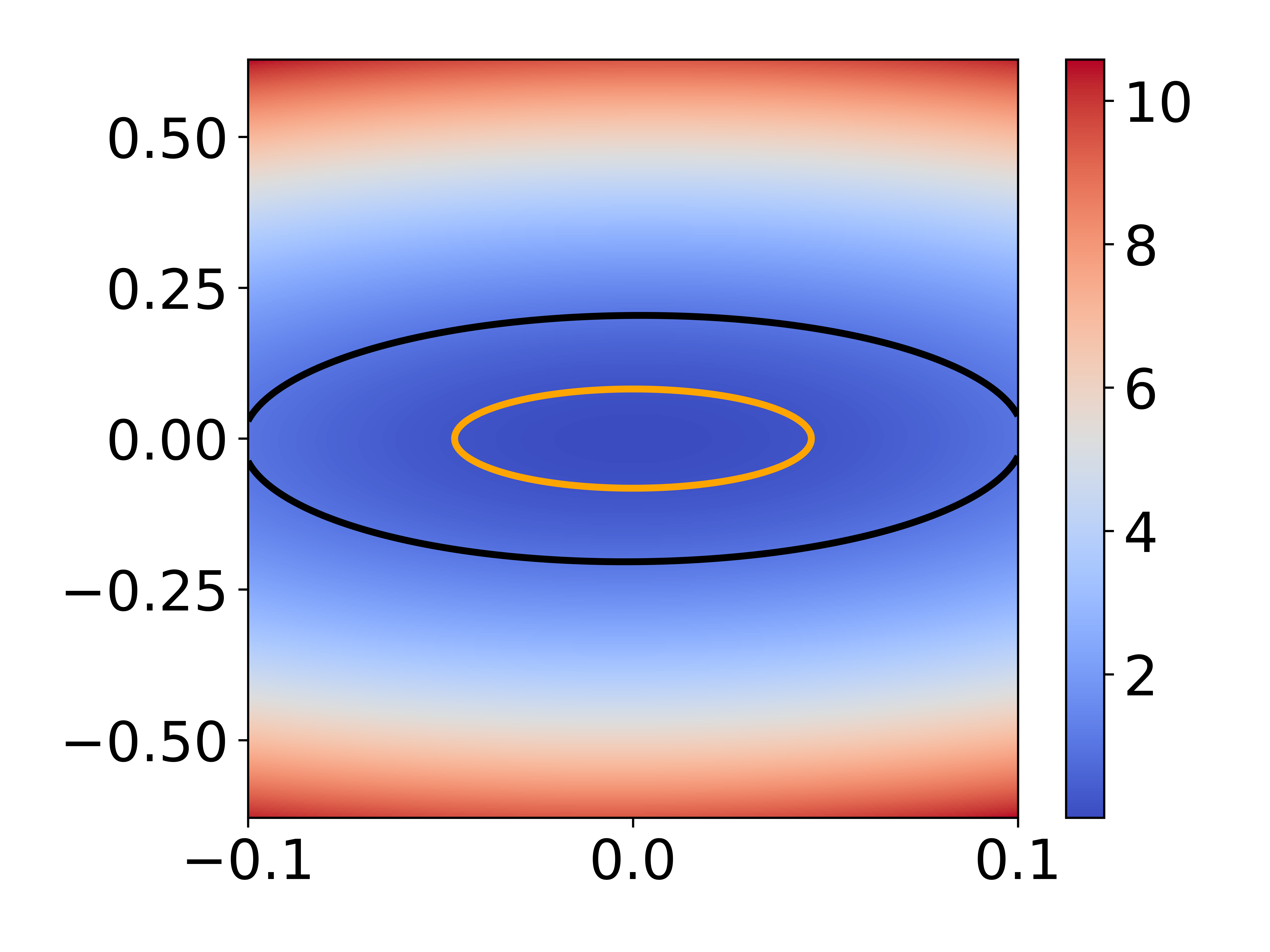}};
            \begin{scope}[x={(image.south east)},y={(image.north west)}]
                \node[below,inner sep=1pt] at (0.5,0) {$y$};
                \node[rotate=90,above,inner sep=1pt] at (0,0.5) {$\theta$};
            \end{scope}
        \end{tikzpicture}%
    }\hfill%
    \subfigure[Projected to $\dot{y}$ and $\dot{\theta}$.]{%
        \begin{tikzpicture}
            \node[anchor=south west,inner sep=0] (image) at (0,0) {\includegraphics[width=0.29\textwidth]{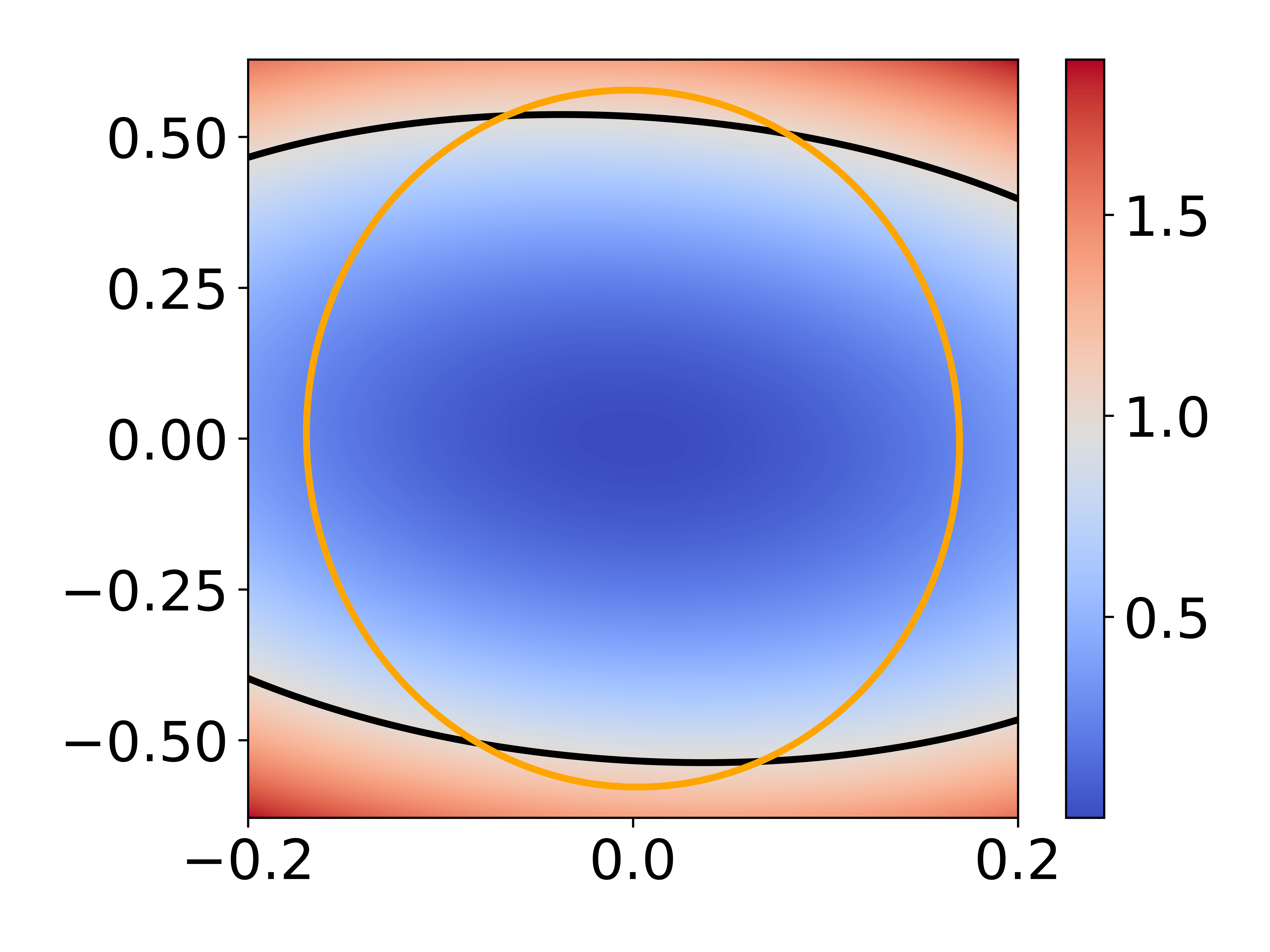}};
            \begin{scope}[x={(image.south east)},y={(image.north west)}]
                \node[below,inner sep=1pt] at (0.5,0) {$\dot{y}$};
                \node[rotate=90,above,inner sep=1pt] at (0,0.5) {$\dot{\theta}$};
            \end{scope}
        \end{tikzpicture}%
    }\hfill%
    \subfigure[Projected to $e_y$ and $e_{\theta}$.]{%
        \begin{tikzpicture}
            \node[anchor=south west,inner sep=0] (image) at (0,0) {\includegraphics[width=0.29\textwidth]{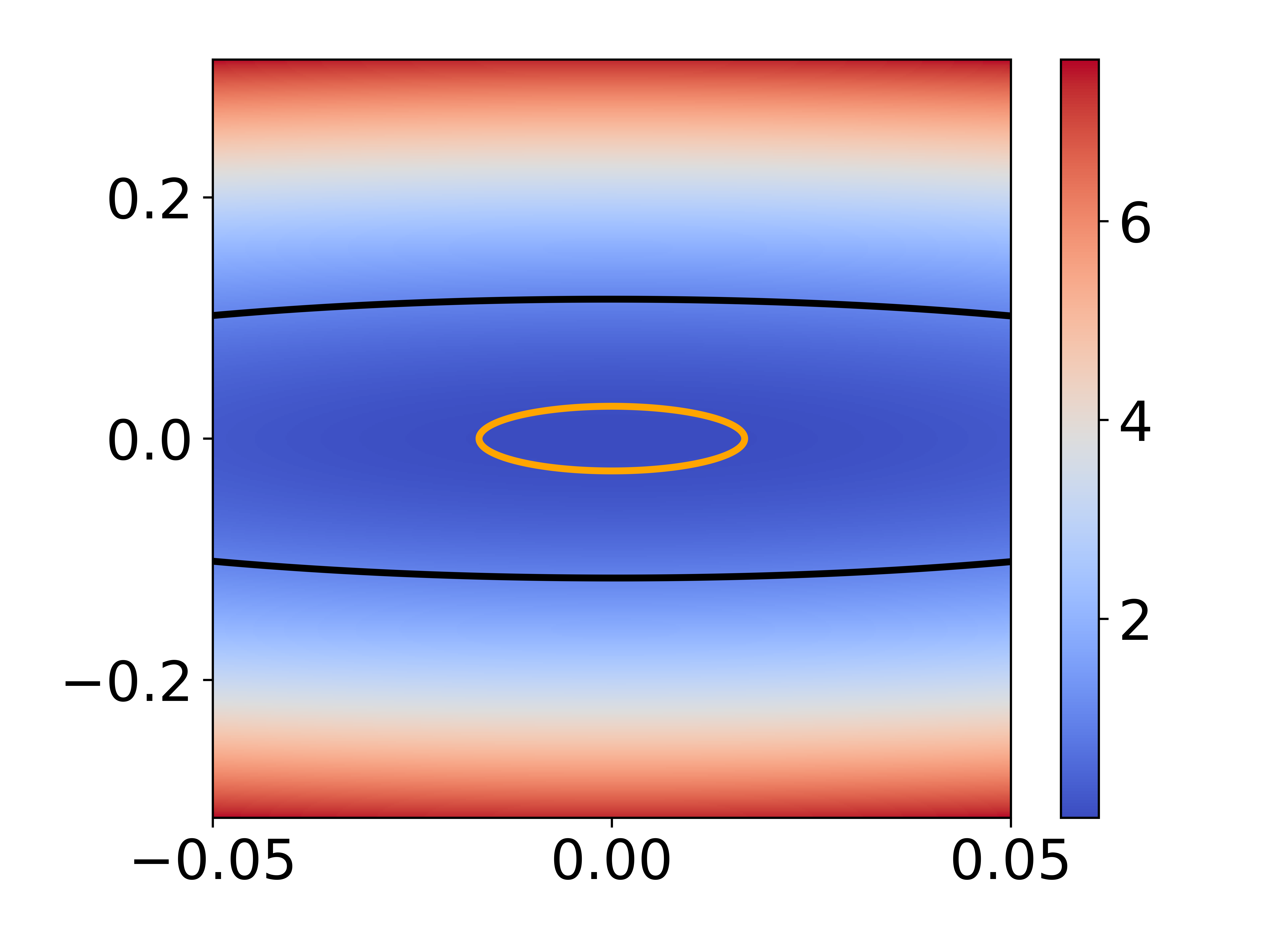}};
            \begin{scope}[x={(image.south east)},y={(image.north west)}]
                \node[below,inner sep=1pt] at (0.5,0) {$e_y$};
                \node[rotate=90,above,inner sep=1pt] at (0,0.5) {$e_{\theta}$};
            \end{scope}
        \end{tikzpicture}%
    }

\caption{
Visualization of the Lyapunov function (color plots) and ROA (contours) compared to \citet{yang2024lyapunov}, on the 2D Quadrotor with output feedback, with three different 2D views.
The system contains 8 states denoted as $ \rvx=[y,\theta,\dot{y},\dot{\theta},e_y,e_\theta,e_{\dot{y}}, e_{\dot{\theta}}]$ (detailed in \Cref{ap:dynamics_output}).
Our method demonstrates a $164\times$ larger ROA (in terms of the volume of ROA on the 8-dimensional input space) compared to \citet{yang2024lyapunov}.
}
\label{fig:roa}
\end{figure}

In \Cref{fig:roa}, we visualize the ROA on the 2D Quadrotor with output feedback, demonstrating much larger ROA compared to   \citet{yang2024lyapunov}

\section{System Dynamics}
\label{ap:system_dynamics}

We adopt system dynamics from existing works~\citep{wu2023neural,yang2024lyapunov}, and we attach them here for completeness. The systems are originally described in continuous time, and since we study discrete-time systems, following \citet{wu2023neural,yang2024lyapunov}, we set a time interval $\Delta t$ with $\Delta t=0.05$ for most systems except for $\Delta t=0.01$ for 2D Quadrotor and output feedback settings.

\subsection{State-Feedback Systems}

\paragraph{Inverted Pendulum.}
The state has 2 dimensions as $\rvx_t=[\theta_t,\dot{\theta}_t]$,
the control input has only 1 dimension as $u_t$,
and the dynamics are:
\begin{equation*}
\rvx_{t+1}=\rvx_t +
\bigg[
\dot{\theta_t},
\frac{-b\dot{\theta_t}+u_t}{ml^2}
+\frac{g\sin\theta_t}{l}
\bigg] \Delta t,
\end{equation*}
where mass is $m=0.15$, length is $l=0.5$, gravity is $g=9.81$, and friction is $b=0.1$.

\paragraph{Path tracking.}

The state has 2 dimensions as $\rvx_t=[e_t,{\theta_e}_t]$,
the control input has only 1 dimension as $u_t$,
and the dynamics are:
\begin{equation*}
\rvx_{t+1}=\rvx_t +
\bigg[
\sin({\theta_e}_t),
\frac{u_t}{l}
- \frac{\cos({\theta_e}_t)}{r - s\sin({\theta_e}_t)}
\bigg] \cdot s \cdot \Delta t,
\end{equation*}
where speed is $s=2.0$, radius is $r=10$, and length is $l=1.0$.

\paragraph{Cart-Pole.}

The state has 4 dimensions as $\rvx_t=[x_t, \dot{x}_t, \theta_t, \dot{\theta}_t]$,
the control input has only 1 dimension as $u_t$,
and the dynamics are:
\begin{equation*}
\rvx_{t+1}=\rvx_t +
{\renewcommand{\arraystretch}{1.35}%
\begin{bmatrix}
\dot{x}_t \\
\frac{
u_t
+ m_p \cdot \sin\theta_t
\cdot (l \cdot \dot{\theta_t}^2
- g \cdot \cos{\theta_t})
}{m_c + m_p \sin^2\theta_t} \\
\dot{\theta}_t \\
\frac{
-u_t \cos{\theta_t}
- m_p l \dot{\theta_t}^2 \cdot \cos{\theta_t} \sin{\theta_t}
+ (m_p+m_c) g \sin{\theta_t}
}{l (m_c + m_p \sin^2\theta_t)}
\end{bmatrix}}
 \Delta t,
\end{equation*}
where length is $l=1.0$, gravity is $g=9.8$,
mass for the cart is $m_c=1.0$,
and mass for the pole is $m_p=0.1$.

\paragraph{PVTOL.}
The state has 6 dimensions as $\rvx_t=[x_t, y_t, \theta_t, \dot{x}_t, \dot{y}_t, \dot{\theta}_t]$,
the control input has 2 dimensions as $\rvu_t$,
and the dynamics are:
$$
\rvx_{t+1}=\rvx_t+
\bigg(
\left[\begin{array}{c}
\dot{x}_t \cos \theta_t- \dot{y}_t \sin \theta_t \\
\dot{x}_t \sin \theta_t+ \dot{y}_t\cos \theta_t \\
\dot{\theta_t} \\
\dot{y}_t \dot{\theta_t}-g \sin \theta_t \\
-\dot{x}_t \dot{\theta_t}-g \cos \theta_t \\
0
\end{array}\right]+\left[\begin{array}{cc}
0 & 0 \\
0 & 0 \\
0 & 0 \\
0 & 0 \\
(1 / m) & (1 / m) \\
l / J & (-l / J)
\end{array}\right] \rvu_t
\bigg) \Delta t,
$$
where $m=4.0$, $l=0.25$, $J=0.0475$, and $g=9.8$.

\paragraph{2D Quadrotor.}
The state has 6 dimensions as $\rvx_t=[x_t, y_t, \theta_t, \dot{x}_t, \dot{y}_t, \dot{\theta}_t]$,
the control input has 2 dimensions as $\rvu_t=[u_{t1},u_{t2}]$,
and the dynamics are:
\begin{equation*}
\rvx_{t+1}=\rvx_t+
\bigg[
\dot{x}_t \Delta t + \frac{\ddot{x_t} (\Delta t)^2}{2},
\dot{y}_t \Delta t + \frac{\ddot{y_t} (\Delta t)^2}{2},
\dot{\theta}_t \Delta t + \frac{\ddot{\theta_t} (\Delta t)^2}{2},
\ddot{x}_t\Delta t,
\ddot{y}_t\Delta t,
\ddot{\theta}_t\Delta t
\bigg],
\end{equation*}
$$ \ddot{x}_t = \frac{-\sin \theta_t ( u_{t1} + u_{t2})}{m}, $$
$$ \ddot{y}_t = \frac{ \cos \theta_t (u_{t1}  + u_{t2}) - g}{m}, $$
$$ \ddot{\theta}_t = \frac{u_{t1} - u_{t2}}{J},$$
where $m=0.486$, $l=0.25$, $J=0.00383$, and $g=9.81$.

\subsection{Output-Feedback Systems}

\label{ap:dynamics_output}

For output-feedback systems, we follow the settings in \citet{yang2024lyapunov} as follows. The state $\rvx_t$ considered in the verification problem is not simply the true state which we denote by $\tilde{\rvx}_t$. Suppose we have the same dynamics with a system output $\rvy_t \in \mathbb{R}^{n_y}$ given by an output map $h(\tilde{\rvx})$ as
\begin{align*}
    \tilde{\rvx}_{t+1} &= f(\tilde{\rvx_t}, \rvu_t), \\
    \rvy_t &= h(\tilde{\rvx_t}).
\end{align*}
We here consider the setting where the controller does not have access to the true state $\tilde{\rvx}_t$, but rather only have the information about the output $\rvy_t$, which could be a subset of $\tilde{\rvx}_t$ or in general any nonlinear function of $\tilde{\rvx}_t$. The state $\hat{\rvx}_t$ is estimated with a dynamic state-observer using a NN $\phi_{\text{obs}}: \mathbb{R}^{n_x}\times \mathbb{R}^{n_y}\to \mathbb{R}^{n_x}$ as
\begin{align*}
    \hat{\rvx}_{t+1} &= f(\hat{\rvx}_t, \rvu_t) + \phi_{\text{obs}}(\hat{\rvx}_t, \rvy_t - h(\hat{\rvx}_t)) - \phi_{\text{obs}}(\hat{\rvx}_t, \vzero_{n_y}).
\end{align*}
The control actions are obtained by
$ \rvu_t = \pi(\hat{\rvx}_t, \rvy_t)$ where $\pi(\cdot,\cdot)$ is a NN-based controller.
Now by considering the augmented system with augmented state $\rvx_t = [\tilde{\rvx}_t, \rve_t]^\top$ where $\rve_t = \hat{\rvx}_t - \tilde{\rvx}_t$, the output-feedback system is the following closed-loop dynamics $\rvx_{t+1} = f_{\text{cl}}(\rvx_t)$ with $f_{\text{cl}}$ being defined as:
\begin{align*}
    f_{\text{cl}}(\rvx_t) &= \begin{bmatrix}
        f(\tilde{\rvx}_t, \pi(\hat{\rvx}_t, h(\tilde{\rvx}_t))) \\
        f(\hat{\rvx}_t, \pi(\hat{\rvx}_t, h(\tilde{\rvx}_t)))+g(\tilde{\rvx}_t, \hat{\rvx}_t) - \tilde{\rvx}_t
    \end{bmatrix},\\
   \text{where} \ \ g(\tilde{\rvx}_t, \hat{\rvx}_t) &= \phi_{\text{obs}}(\hat{\rvx}_t, h(\tilde{\rvx}_t) - h(\hat{\rvx}_t)) -\phi_{\text{obs}}(\hat{\rvx}_t, \vzero_{n_y}).
\end{align*}
We follow \citet{yang2024lyapunov} to consider the following two output-feedback dynamics.

\paragraph{Inverted Pendulum with output feedback.} The system dynamics are the same as the state-feedback setting with output map $h(\rvx_t) = \theta_t$.

\paragraph{2D Quadrotor with output feedback.}
There is a $4$-dimensional state $\tilde{\rvx}= [p, \theta, \dot{p}, \dot{\theta}]$ with two control inputs $\rvu = [u_1, u_2]$:
\begin{align*}
    \tilde{\rvx}_{t+1} &= \tilde{\rvx}_t + \begin{bmatrix}
        \dot{p}_t \\
        \dot{\theta}_t \\
        \frac{1}{m} \cos(\theta_t)(u_1+u_2) - g \\
        \frac{l}{I} (u_1- u_2)
    \end{bmatrix}\Delta_t
\end{align*}
where $m=0.486, l=0.25, I=0.00383, g=9.81, \Delta_t=0.01$. This system obtains observations from a lidar sensor. Let $\alpha_{\max} = 0.149\pi$ and $\alpha_i$ be $4$ angles evenly spaced in $[-\alpha_{\max}, \alpha_{\max}]$. Let $H=5$ and $h_0=1$, then we define
$$h(x) = [h_1(x), \cdots h_4(x)]^\top
\quad\text{s.t.}
\enskip h_i(x) = \text{clamp}(\frac{p+h_0}{\cos(\theta - \alpha_i)}, 0, H),$$
as the output map.

\end{document}